%% file: main.tex
\documentclass[10pt,twocolumn,letterpaper]{article}

\usepackage[review]{cvpr}      % To produce the REVIEW version

\usepackage{graphicx}
\usepackage{amsmath}
\usepackage{amssymb}
\usepackage{booktabs}
\usepackage{color}
\usepackage{float}
\usepackage{bbding}
\usepackage{mwe}

\definecolor{cvprblue}{rgb}{0.21,0.49,0.74}
\usepackage[pagebackref,breaklinks,colorlinks,allcolors=cvprblue]{hyperref}
\usepackage{caption}
\usepackage{xcolor} % 加载颜色包
\usepackage{multirow}
\usepackage{colortbl}
\usepackage{pifont}
\usepackage{booktabs}

\newcommand{\cmark}{\ding{51}} % ✓
\newcommand{\xmark}{\ding{55}} % ✗
\definecolor{mycamored}{RGB}{255,35,35} % 定义自定义颜色
\definecolor{mycamoblue}{RGB}{68,114,196} % 定义自定义颜色
\definecolor{mycamoyellow}{RGB}{255,192,0} % 定义自定义颜色

\usepackage[pagebackref,breaklinks,colorlinks]{hyperref}

\title{GenCAMO: Scene-Graph Contextual Decoupling for Environment-aware and Mask-free Camouflage Image-Dense Annotation Generation}
\author{
    Chenglizhao Chen$^{1}$, Shaojiang Yuan$^{1}$, Xiaoxue Lu$^{1}$, Mengke Song$^{1}$, Jia Song$^{1}$,\\ Zhenyu Wu$^{2}$, Wenfeng Song$^{3}$, Shuai Li$^{4}$\\
    $^1$China University of Petroleum (East China)\\
    $^2$Southwest Jiaotong University\\
    $^3$Beijing Information Science and Technology University\\
    $^4$Beihang University\\
}

\begin{document}

\twocolumn[{
	\maketitle
	\vspace{-10mm}
	\begin{center}
		\centering
		\includegraphics[width=\linewidth]{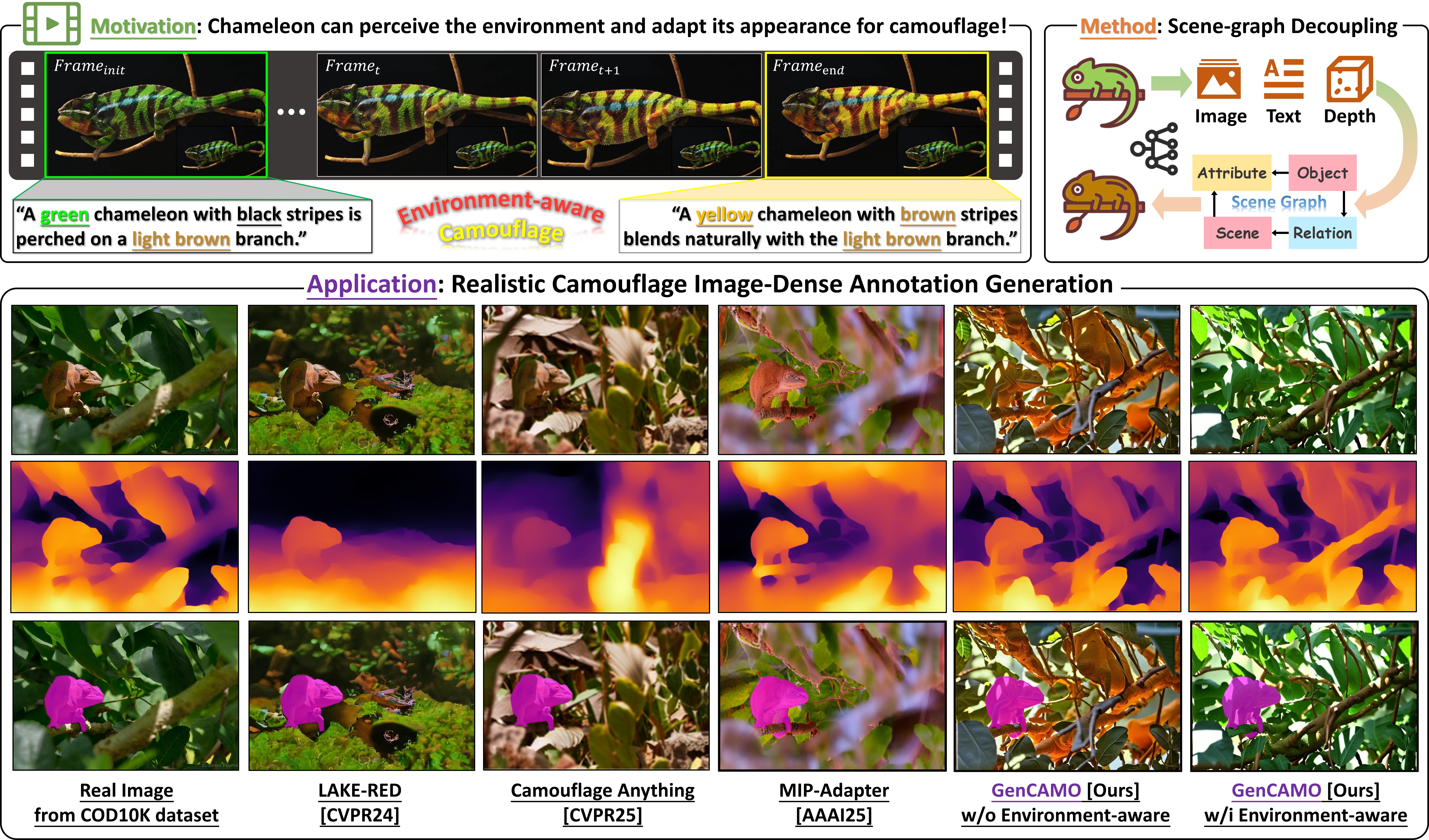}
		\vspace{-7mm}
		\captionof{figure}{\textbf{Motivation, Method, and Application of this work.} Inspired by the environment-aware camouflage ability\cite{teyssier2015photonic} of chameleons, \textbf{\textit{GenCAMO}} is a mask-free generative framework that takes image–text–depth 
			conditions as input and produces realistic and context-adaptive camouflage images together with 
			their depth and mask annotations. These outputs are guided by a 
			scene-graph decoupling mechanism that separates object attributes, 
			relations, and environmental cues to achieve controllable generation.
		}
		\vspace{1mm}
		\label{fig:motivation}
	\end{center}
}]

\input{sec/0_abstract}    
\input{sec/1_intro}

\input{sec/2_formatting}
\input{sec/3_finalcopy}

\input{sec/X_suppl}
{
    \small
    \bibliographystyle{ieeenat_fullname}
    \bibliography{main}
}

\end{document}

%% file: sec/0_abstract.tex
\begin{abstract}
Conceal dense prediction (CDP), especially RGB-D camouflage object detection and open-vocabulary camouflage object segmentation, plays a crucial role in advancing the understanding and reasoning of complex camouflage scenes.
However, high-quality and large-scale camouflage datasets with dense annotation remains scarce because of expensive data collection and labeling costs.
To address this challenge, we explore leveraging generative models to synthesize realistic camouflage image-dense data for training CDP models with fine-grained representations, prior knowledge, and auxiliary reasoning. 
Concretely, our contribution are threefold: (i) we introduce GenCAMO-DB, a large-scale camouflage dataset with multi-modal annotations, including depth maps, scene graphs, attribute descriptions, and text prompts; (ii) we present GenCAMO, an environment-aware and mask-free generative framework that produces high-fidelity camouflage image–dense annotations; (iii) extensive experiments across multiple modalities demonstrate that GenCAMO significantly improves dense prediction performance on complex camouflage scene by providing high-quality synthetic data.
\end{abstract}

%% file: sec/1_intro.tex
\section{Introduction}
\label{sec:intro}

\textbf{Background.} Camouflaged object detection (COD)\cite{fan2020camouflaged, yin2024camoformer,pang2024zoomnext} has achieved remarkable success by leveraging extensive manual image-mask annotations, playing a crucial role in various real-world applications such as agriculture\cite{yang2024plantcamo}, industrial inspection\cite{kumar2008computer}, and ecological monitoring\cite{rustia2020application}. 
However, developing models for Concealed Dense Prediction (CDP)\cite{zhao2025deep}, including depth-guided camouflage object detection (RGB-D COD)\cite{wang2024depth} and open-vocabulary camouflaged object segmentation (OVCOS)\cite{pang2024open}, remains challenging due to dense-data scarcity, modality complexity, and the high cost of dense manual annotations.
These limitations severely hinder the advancement of dense prediction techniques in camouflage scene.

\noindent
\textbf{Existing approaches and challenges.} In the field of camouflage image generation, mainstream methods\cite{zhao2024lake,das2025camouflage,chen2025foreground,zhang2023adding} are based on foreground object image outpainting. These approaches take camouflaged foreground objects and camouflage masks as inputs, synthesizing visually consistent camouflage images by adjusting the appearance and texture of the background.
However, these methods rely on manually annotated foreground masks, which increases the labeling workload. Additionally, without realistic spatial layout modeling or fine-grained contextual semantics, outpainting-based synthesis often produces distorted depth maps and background regions dominated by foreground appearance (Fig.~\ref{fig:motivation}). As a result, the generated camouflage objects fail to perceive and adapt to their environments, ultimately limiting performance in downstream tasks, especially dense prediction.

\begin{table*}
	\centering
	\caption{Comparison of camouflage object detection (COD), camouflage dense prediction (CPD), camouflage image generation (CIG) and our camouflage image-dense annotation generation (CIDG) datasets. FG Attr. represents fine-grained attributes, SG denotes scene graphs, and Anno. indicates annotations.}
	\label{tab:datasets}
		\resizebox{0.9\textwidth}{!}{
	\begin{tabular}{ccccccccccccccc} 
		\toprule
		\multirow{2}{*}{\textbf{Domain}} & \multirow{2}{*}{} & 
		\multirow{2}{*}{\textbf{Dataset}} & \multirow{2}{*}{} & 
		\multirow{2}{*}{\textbf{Year}} & \multirow{2}{*}{} & 
		\multicolumn{5}{c}{\textbf{Modalities}} &  & 
		\multicolumn{3}{c}{\textbf{Number}}                   \\ 
		\cline{7-11}\cline{13-15}
		& & & & & &
		\textbf{Image} & \textbf{Text} & \textbf{Depth} & 
		\textbf{FG Attr.} & \textbf{SG} & & 
		\textbf{Samples} & \textbf{Words}  & \textbf{Anno.}   \\ 
		\hline
		
		% ================= COD =================
		\multirow{5}{*}{COD} & \multirow{5}{*}{} 
		& CHAMELEON\cite{skurowski2018animal} & & 2018 & 
		& \cmark & \xmark & \xmark & \xmark & \xmark & 
		& 76   & -   & - \\
		
		& & CAMO\cite{le2019anabranch} & & 2019 & 
		& \cmark & \xmark & \xmark & \xmark & \xmark & 
		& 1.2K & -   & - \\
		
		& & COD10K\cite{fan2020camouflaged} & & 2020 & 
		& \cmark & \xmark & \xmark & \xmark & \xmark & 
		& 5K   & -   & - \\
		
		& & NC4K\cite{lv2021simultaneously} & & 2021 & 
		& \cmark & \xmark & \xmark & \xmark & \xmark & 
		& 4.1K & -   & - \\
		
		& & USC12K\cite{zhou2025rethinking} & & 2025 & 
		& \cmark & \xmark & \xmark & \xmark & \xmark & 
		& 12K  & -   & - \\ 
		\hline
		
		% ================= CPD =================
		\multirow{3}{*}{CPD} & 
		& CODD\cite{zhang2024new} & & 2024 & 
		& \cmark & \xmark & \cmark & \xmark & \xmark & 
		& 455  & -   & 455 \\
		
		& & ACOD12K\cite{wang2024depth} & & 2024 &
		& \cmark & \xmark & \cmark & \xmark & \xmark & 
		& 6K   & -   & 6K \\
		
		& & OVCAMO\cite{pang2024open} & & 2024 & 
		& \cmark & \cmark & \cmark & \xmark & \xmark & 
		& 12K  & 12K & - \\ 
		\hline
		
		% ================= CIG =================
		\multirow{2}{*}{CIG} & 
		& LCGNET\cite{li2022location} & & 2022 & 
		& \cmark & \xmark & \xmark & \xmark & \xmark & 
		& 5K  & - & - \\
		
		& & LAKE-RED\cite{zhao2024lake} & & 2024 & 
		& \cmark & \xmark & \xmark & \xmark & \xmark & 
		& 17K & - & - \\ 
		\hline
		
		% ================= CIDG =================
		\rowcolor[rgb]{0.933,0.918,0.949} 
		CIDG & & GenCAMO-DB & & Ours &
		& \cmark & \cmark & \cmark & \cmark & \cmark &
		& \textbf{34.2K} & \textbf{612.5K} & \textbf{102.6K}  \\
		\bottomrule
	\end{tabular}
}
\end{table*}

\noindent
\textbf{Motivation and contributions.} Based on these observations, we propose an environment-aware and mask-free camouflage image–dense annotation generation framework, \textbf{GenCAMO}, which can jointly leverage scene semantics, spatial depth, and contextual relationships to achieve fine-grained, geometry-consistent camouflage generation.
To alleviate data scarcity, we construct \textbf{GenCAMO-DB}, a high-quality and large-scale dataset containing nearly 34,200 camouflage-related images with multimodal dense annotations. It integrates data from both general and camouflage-specific sources and provides rich labels, including depth maps, fine-grained attributes, text prompts, and scene graphs, to support environment-aware camouflage generation.

For mask-free camouflage generation, we propose GenCAMO, a reference-guided and depth-conditioned text-to-image framework capable of synthesizing camouflage dense data without manual masks. The main challenge is representing concealed objects in complex scenes under mask-free conditions. To overcome this, we introduce a scene-graph contextual decoupling mechanism that separates spatial layouts and object attributes for fine-grained controllable generation. GenCAMO further incorporates two key modules: (i) Depth Layout Coherence Guided ControlNet that reinforces object–background spatial consistency, and (ii) Attribute-aware Mask Attention, which injects scene-graph-derived attribute cues to improve appearance adaptation and cross-modal alignment. Finally, decoder features are shared across image, depth, and mask decoders, enabling fully mask-free generation of camouflage image–dense annotations to support concealed dense prediction tasks.

Overall, our contributions can be summarized as follows:
(i) we explore leveraging reference-guided text-to-image generative model for camouflage image-dense annotation generation without mask condition, facilitating training conceal dense prediction models for various camouflage scene;
(ii) we construct GenCAMO-DB, a large-scale camouflage dataset with multi-modal annotations, including depth maps, scene graphs, fine-grained attribute descriptions, and text prompts, serves as a solid basis for camouflage generative modeling;
(iii) we propose GenCAMO, an environment-aware and mask-free generative framework that produces high-fidelity camouflage image–dense annotations;
(iv) we conduct extensive experiments across various conceal dense predict task, demonstrating that our GenCAMO can enhance the robustness of camouflage scene understanding models in unannotated field.
To the best of our knowledge, this work presents the first open-source dataset curated for camouflage image–dense annotation generation, and the first text-to-image generative framework specifically designed for camouflage-mask-free condition.

\section{Related Work}
\label{sec:Related Work}
%\paragraph{Conceal Dense Prediction} xxxx
\textbf{Synthetic Camouflaged Dataset Generation.} Early works \cite{chu2010camouflage} achieve camouflage image generation by adjusting backgrounds to match fixed foregrounds in color and texture. Recent methods introduce GANs \cite{goodfellow2020generative}, diffusion models \cite{zhao2024lake}, or outpainting ControlNets \cite{das2025camouflage} to improve realism, yet still rely on manual annotated foreground mask. In contrast, our approach removes the need for mask supervision by using reference-guided diffusion and scene-graph contextual cues to generate camouflage images and dense annotations in a fully mask-free manner.

\noindent
\textbf{Text-to-Image Generation.} Recent text-to-image methods enable controllable synthesis for general dataset construction and object segmentation\cite{wu2023diffumask}. 
Approaches like DatasetDiffusion\cite{nguyen2023dataset} and MaskFactory\cite{qian2024maskfactory} utilize generic prompts or random sampling, whereas reference-based methods such as GLIGEN\cite{li2023gligen}, DreamBooth\cite{ruiz2023dreambooth}, and MS-Diffusion\cite{wang2024ms} enhance alignment via visual examples. 
However, they cannot capture adaptive camouflage cues, fail to model foreground–background relations, and still rely on mask supervision, making them unsuitable for dense annotation synthesis in mask-scarce settings.

%-------------------------------------------------------------------------

%% file: sec/2_formatting.tex
\begin{figure}[t]
	\centering
	\includegraphics[width=0.9\linewidth]{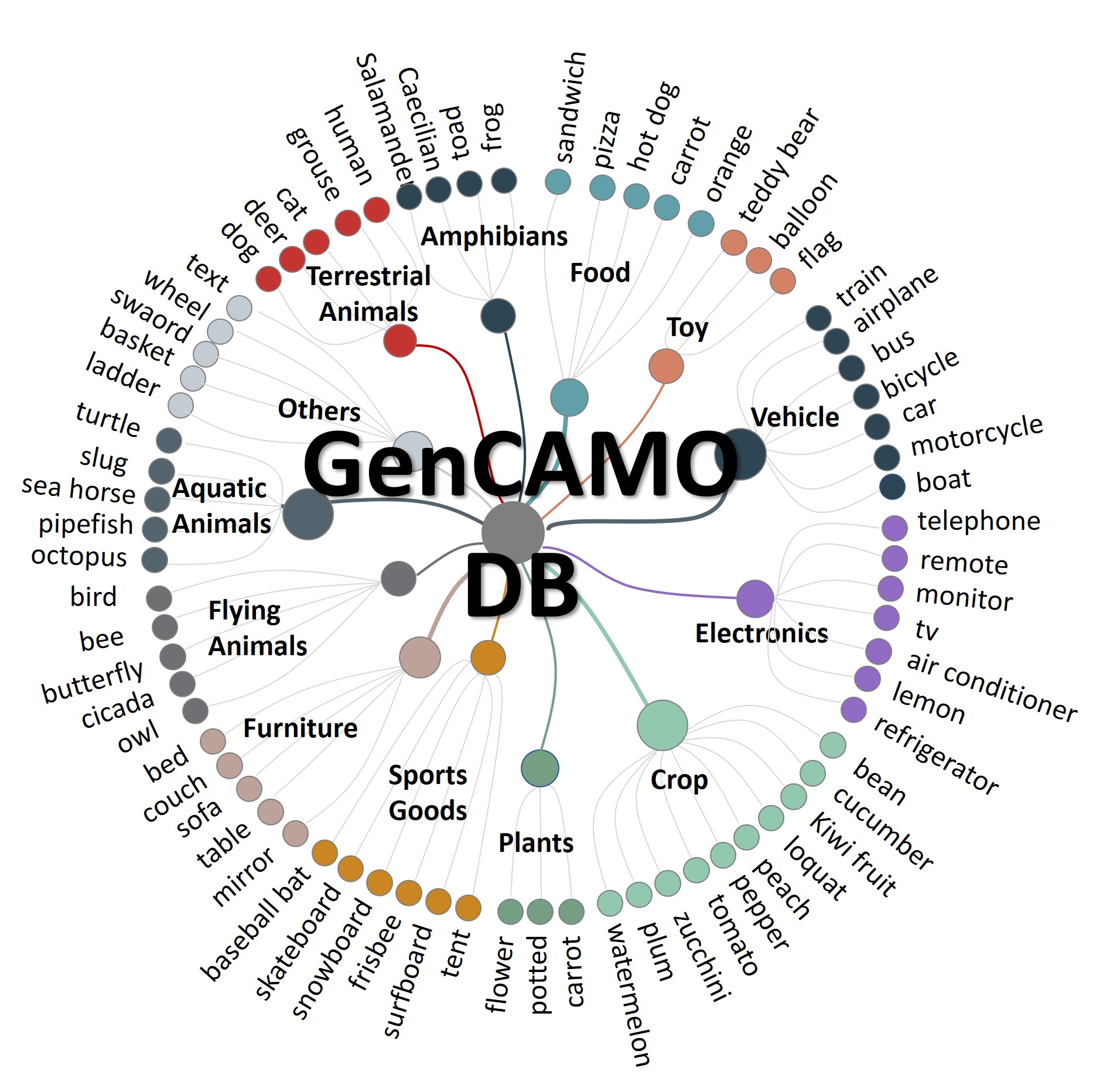}
	\caption{Illustration of the semantic concepts distribution for
		the concealed, salient and general categories in our GenCAMO-DB.}
	\label{fig:data_statistic}
\end{figure}

\begin{figure}[t]
	\centering
	\includegraphics[width=0.9\linewidth]{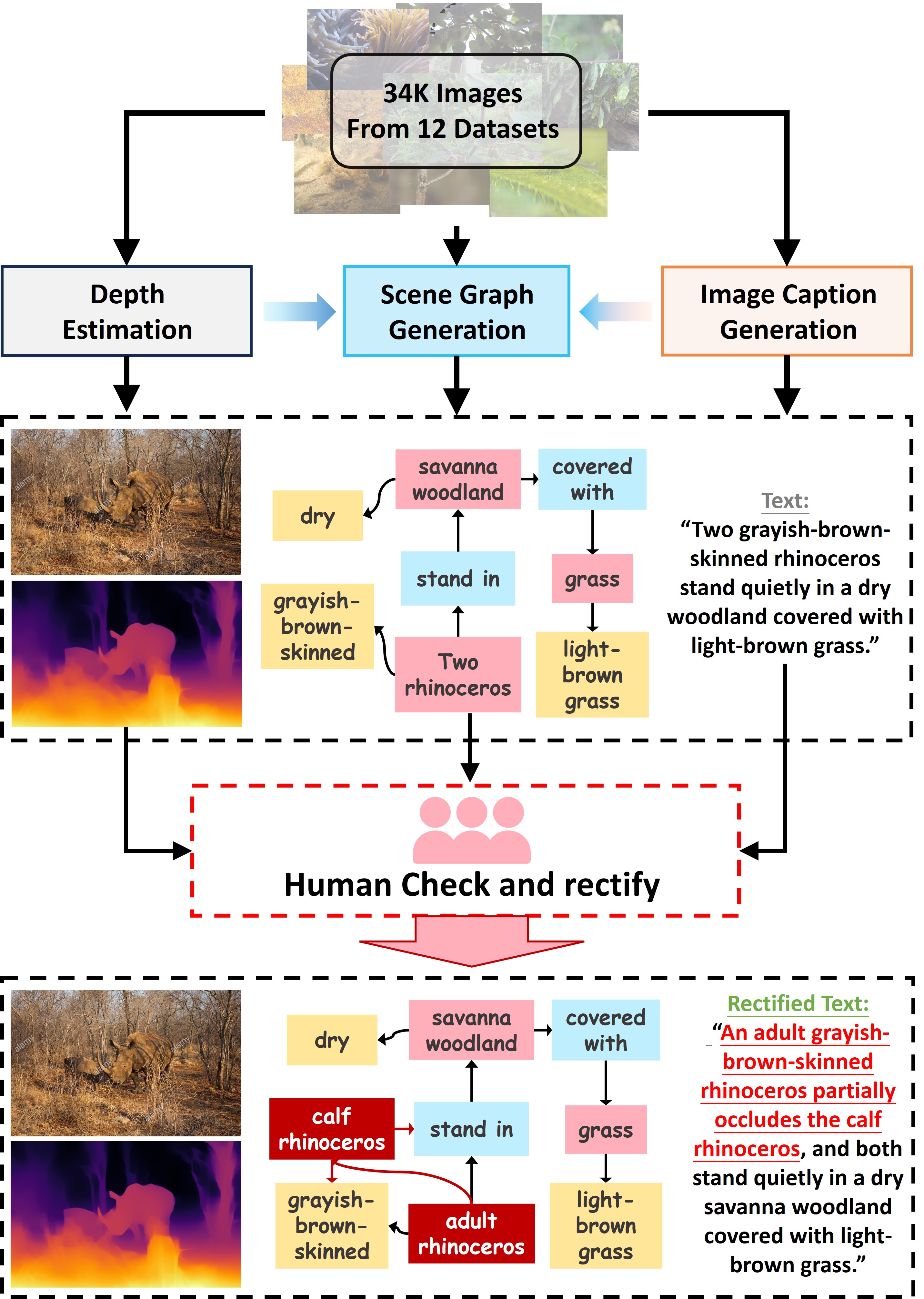}
	\caption{Overview of our dataset construction pipeline. Depth maps, scene graphs, and captions are automatically generated for 34K images, followed by human verification and refinement.}
	\label{fig:data_construc}
\end{figure}

\section{GenCAMO-DB Dataset}
The scarcity of camouflage images with high-quality dense annotations poses significant challenges for training generative models. To address this issue, we introduce \textbf{GenCAMO-DB}, a large-scale camouflage image–dense annotation dataset that provides concealment-oriented text prompts, accurate depth maps, and structured scene-graph representations across diverse scenes. As illustrated in Fig.~\ref{fig:data_statistic}, the dataset spans a wide range of domains, including natural, household, agricultural, and industrial environments.
In the following sections, we describe the dataset construction pipeline and present detailed statistics and analyses of our dataset.

\begin{figure*}[t]
	\centering
	\includegraphics[width=\linewidth]{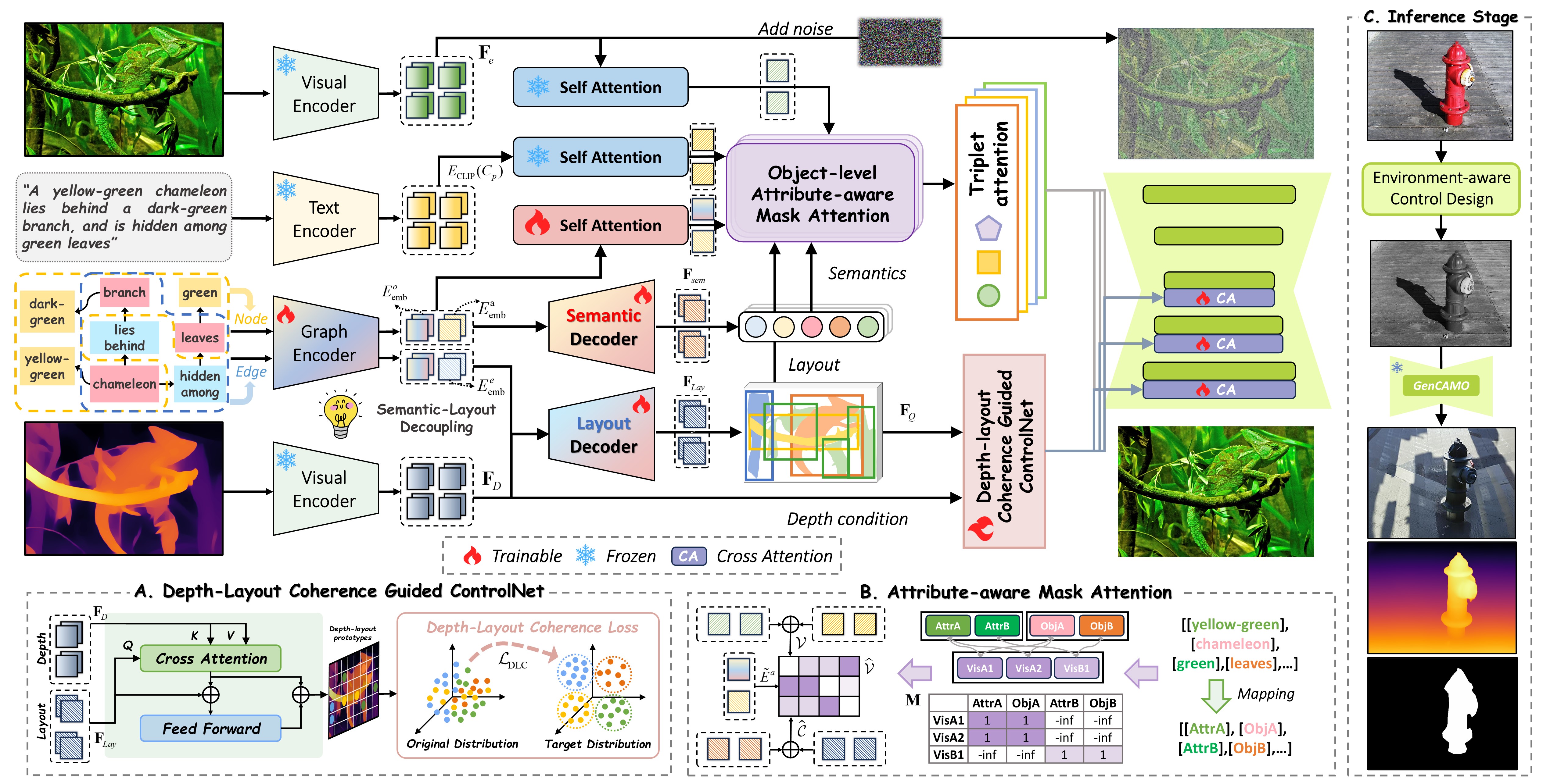}
	\caption{Overview of the proposed method framework. 
		GenCAMO integrates visual, textual, and scene-graph cues through 
		semantic–layout decoupling, depth–layout coherence guidance, and 
		attribute-aware mask attention to generate context-adaptive camouflage 
		images with corresponding depth and mask annotations.}
	\label{fig:model_1}
\end{figure*}

\subsection{Data collection}
Owing to the mask-free paradigm adopted in our generative framework, GenCAMO-DB eliminates the dependency on precise pixel-level camouflage mask annotations, which are often labor-intensive, ambiguous, and scene-specific. 
This property enables us to explore a much wider range of potential camouflage scenarios without being constrained by annotation availability.
As show in Fig.\ref{fig:data_construc}, to ensure diversity and completeness, we construct GenCAMO-DB from three sources: (i) open-domain datasets with scene-graph annotations (e.g., COCO-Stuff, Visual Genome), from which we manually select camouflage-like scenes; (ii) camouflaged image datasets, for which we generate depth maps, scene graphs, and text prompts through a semi-automatic pipeline; and (iii) SOD and SEG images from LAKERED, which we extend with corresponding dense annotations for compatibility with existing benchmarks.
\subsection{Semi-Automatic annotations}
To build a comprehensive multi-modal camouflage dataset, we process 34,200 images from 12 open-source datasets through a unified pipeline that generates depth maps, scene graphs, and captions. Depths are produced by Depth Anything \cite{yang2024depth}, and scene graphs are generated by Universal SG \cite{wu2025universal} and refined with camouflage-specific textual cues. Captions are created using GPT-4o \cite{hurst2024gpt}. All modalities undergo human verification for depth consistency, scene-graph correctness, and attribute alignment, with 5–10 minutes spent per image; samples failing camouflage-likelihood or cross-modal checks are re-annotated to ensure high-quality, coherent results.

\section{Methodology}
\label{sec:Method}

\textbf{Preliminary: Camouflage Scene Graph Representation.}
As shown in Fig.\ref{fig:model_1}, the scene graph $G = (O, E)$ defines a structured abstraction of the scene. \textbf{\textcolor{mycamored}{Nodes}} $O = \{o_i\}_{i=1}^{N_o}$ correspond to the $N_o$ \textbf{\textcolor{mycamored}{object}} entities in the scene, such as ``chameleon" and ``branch", whereas \textbf{\textcolor{mycamoblue}{edges}} $E = \{e_{ij}\}_{1 \le i,j \le N_o,\, i \ne j}$ capture their pairwise relationships. For instance, the edge between node ``chameleon" and ``branch" is ``lies behind”.
In order to model the low-level appearance and texture cues crucial for camouflage, we further incorporate a set of conceal \textbf{\textcolor{mycamoyellow}{attributes}} $A = \{a_i\}_{i=1}^{N_o}$ describing color, pattern, and material properties for each object.
In practice, the node $O = \{o_i\}_{i=1}^{N_o}$ and the quintuples $\mathcal{T} = \{t_{ij} = (a_i, o_i, e_{ij}, o_j, a_j)\}_{1 \le i,j \le N_o,\, i \ne j}$ represent connections from object $o_i$ with attribute $a_i$ to object $o_j$ with attribute $a_j$. 
These quintuples serve as inputs for graph convolutional networks (GCNs) to perform relational reasoning.
Moreover, objects, attributes, and relations are converted into learnable embeddings using embedding layers denoted as $E^{o}_{\text{emb}}$, $E^{a}_{\text{emb}}$, and $E^{e}_{\text{emb}}$.

% Eo+Ea->semantic, Eo+Ee->layout
%------------------------------------------------------------------------
\noindent
\textbf{Method.} As illustrated in Fig.~\ref{fig:model_1}, we propose \textbf{GenCAMO}, a multi-condition guided framework for camouflage image–dense annotation generation. GenCAMO consists of three core components: (i) a Depth–Layout Coherence Guided ControlNet (DLCG-ControlNet) that fuses scene-graph layouts with depth cues for geometry-consistent features; (ii) an Attribute-aware Mask Attention (AMA) module that aligns object and attribute relations in the diffusion process; and (iii) a unified generation module that synthesizes controllable camouflage images and jointly decodes images, masks, and depth maps.

\subsection{Depth Layout Coherence Guided ControlNet}
A key challenge in depth-conditioned ControlNet is capturing object-level relations in camouflage scenes, which we address by aligning depth features with textual prompts through scene-graph–based layout embeddings.
Given an input depth condition $C_d$, the corresponding scene graph node feature $E^{o}_{\text{emb}}$, and edge feature $E^{e}_{\text{emb}}$, the depth embedding is extracted using a visual encoder:
\begin{equation}
\mathbf{F}_D = \mathrm{VisualEnc}(C_d),
\label{eq:depthenc}
\end{equation}
while the layout embedding is obtained by decoding object and relation embeddings from the scene graph:
\begin{equation}
\mathbf{F}_{\text{lay}} = \mathrm{LayoutDec}(E^{o}_{\text{emb}} \odot E^{e}_{\text{emb}}),
\label{eq:layoutdec}
\end{equation}
where $\mathbf{F}_D, \mathbf{F}_{\text{lay}} \in \mathbb{R}^{N \times C}$,
$N$ is the number of tokens and $C$ is the feature dimension.
To inject layout information into the depth branch, we fuse the two
features by a learnable linear projection:
\begin{equation}
\mathbf{F}_Q = \mathbf{F}_D + \mathbf{F}_{\text{lay}} W^{L},
\label{eq:fq_fusion}
\end{equation}
where $W^{L} \in \mathbb{R}^{C \times C}$ aligns the layout features
to the depth feature space, and $\mathbf{F}_Q \in \mathbb{R}^{N \times C}$
is the depth--layout fused representation.
To summarize the fused depth-layout features, we introduce $M$ learnable
tokens $\mathbf{T}=\{t_1,\ldots,t_M\}$ and apply cross-attention between
$\mathbf{T}$ and the fused representation $\mathbf{F}_Q$. The resulting
tokens form a compact prototype set
\begin{equation}
\mathbf{P} = \{p_1,\ldots,p_M\}, \quad p_m \in \mathbb{R}^{C},
\end{equation}
which encodes depth--layout priors and provides structural guidance for
the ControlNet branch.
\paragraph{Depth-layout coherence loss.}
To encourage the fused depth features to form compact, object-wise
clusters that are consistent with the scene layout, we define a
depth--layout coherence loss.
For each fused token $\mathbf{F}_Q(i)$, we compute its distance to the
nearest prototype:
\begin{equation}
d_i = \min_{m \in \{1,\dots,M\}}
\big( 1 - \mathcal{S}(\mathbf{F}_Q(i), p_m) \big),
\label{eq:dlc_di}
\end{equation}
where $\mathcal{S}(\cdot,\cdot)$ denotes the cosine similarity.
The overall coherence loss is then written as
\begin{equation}
\mathcal{L}_{\text{DLC}}
=
\frac{1}{N} \sum_{i=1}^{N} d_i,
\label{eq:dlc_loss}
\end{equation}

\subsection{Attribute-aware Mask Attention}
To better align the complex camouflage visual-text information, we obtained the scene-graph semantics embedding by decoding object and attribute embeddings from the scene graph:
\begin{equation}
\mathbf{F}_{\text{sem}} = \mathrm{SemanticsDec}(E^{o}_{\text{emb}} \odot E^{a}_{\text{emb}}),
\label{eq:semanticsdec}
\end{equation}
We integrate the spatial layout feature $\mathbf{F}_{\text{lay}}$ and interactive semantics $\mathbf{F}_{\text{sem}}$ to obtain the object-level embedding.
\begin{equation}
\hat{c}_i =
\begin{cases}
\mathbf{F}_{\text{lay}}^{(i)} \odot \mathbf{F}_{\text{sem}}^{(i)}, & i \le N_o, \\[4pt]
\hat{c}_{\text{null}}, & \text{otherwise,}
\end{cases}
\label{eq:obj_fuse}
\end{equation}
For each object $i$, the fused embedding $\hat{c}_i$ jointly encodes the geometric layout cues (e.g., position and scale) and the relational semantics within the scene graph.
To handle varying numbers of objects, we further introduce a learnable null embedding $\hat{c}_{\text{null}}$ and pad the embedding set to a fixed length $N_{\max}$.
We employ self-attention to process the text feature $E_{\text{CLIP}}(C_p)$, 
the image visual feature $\mathbf{F}_e$, and the raw attribute feature 
$E^{a}_{\text{emb}}$, producing the self-attended attribute tokens 
$\tilde{E}^{a}$. The self-attended text and image features are fused into 
the visual token set $\mathcal{V}$. Given the fused object embedding 
$\hat{\mathcal{C}}$, we integrate $\mathcal{V}$, $\hat{\mathcal{C}}$, 
and $\tilde{E}^{a}$ into the attribute-aware mask attention (AMA) module. 
Following compositional masked attention, the AMA layer is formulated as:
\begin{equation}
\hat{\mathcal{V}} = AMA\big([\mathcal{V} \oplus \hat{\mathcal{C}} \oplus 
\tilde{E}^{a}], \mathbf{M}\big)[:N_v].
\end{equation}
where $\mathbf{M}$ is the attribute-aware attention mask defined as
\begin{equation}
M_{i,j} =
\begin{cases}
1, & \text{if } (i,j) \text{ fall into the same entity,} \\[4pt]
-\infty, & \text{otherwise.}
\end{cases}
\label{eq:mask}
\end{equation}
This design ensures that each visual token only attends to its relevant object and attribute embeddings, avoiding incorrect cross-object interactions.

\begin{table*}
	\centering
	\caption{Quantitative Performance. 
		The performance of the proposed GenCAMO method is 
		quantitatively evaluated against state-of-the-art (SOTA) techniques. 
		$\mathcal{F}$ denotes using only the foreground input, while 
		$\mathcal{F}+\mathcal{B}$ denotes using both foreground and background. 
		$\mathcal{I}$, $\mathcal{T}$, and $\mathcal{D}$ represent the image, 
		text, and depth-map conditions, respectively.}
	\renewcommand{\arraystretch}{1.3}
	\resizebox{\linewidth}{!}{
		\begin{tabular}{ccccccccccc} 
			\toprule     % 粗线
			\multirow{2}{*}{}                                                                                               & \multirow{2}{*}{Methods (Venue)}         & \multirow{2}{*}{Input}                                              & \multicolumn{2}{c}{\textbf{Camouflaged Objects}}                                       & \multicolumn{2}{c}{\textbf{Salient Objects}}                                           & \multicolumn{2}{c}{\textbf{General Objects}}                                           & \multicolumn{2}{c}{\textbf{Overall}}                                                    \\ 
			\cline{4-11}
			&                                          &                                                                     & \textbf{FID$\downarrow$}                  & \textbf{KID$\downarrow$}                   & \textbf{FID$\downarrow$}                  & \textbf{KID$\downarrow$}                   & \textbf{FID$\downarrow$}                  & \textbf{KID$\downarrow$}                   & \textbf{FID$\downarrow$}                  & \textbf{KID$\downarrow$}                    \\ 
			\hline
			\multirow{3}{*}{\begin{tabular}[c]{@{}c@{}}\textbf{\textit{Image}}\\\textbf{\textit{Blendinge~}}\end{tabular}}  &  CI (\textbf{TOG 2010})                   & $\mathcal{F}+\mathcal{B}$                                           & 124.49                                    & 0.0662                                     & 136.30                                    & 0.0738                                     & 137.19                                    & 0.0713                                     & 128.51                                    & 0.0693                                      \\
			&  DCI (\textbf{AAAI 2020})                 & $\mathcal{F}+\mathcal{B}$                                           & 130.21                                    & 0.0689                                     & 134.92                                    & 0.0665                                     & 137.99                                    & 0.0690                                     & 130.52                                    & 0.0675                                      \\
			& LCGNet (\textbf{TMM 2023})~              & $\mathcal{F}+\mathcal{B}$                                           & 129.80                                    & 0.0504                                     & 136.24                                    & 0.0597                                     & 132.64                                    & 0.0548                                     & 129.88                                    & 0.0550                                      \\ 
			\hline
			\multirow{3}{*}{\begin{tabular}[c]{@{}c@{}}\textit{\textbf{Image }}\\\textit{\textbf{Inpainting}}\end{tabular}} &  LDM (\textbf{CVPR 2022})~                & $\mathcal{F}$                                                       & 58.65                                     & 0.0380                                     & 107.38                                    & 0.0524                                     & 129.04                                    & 0.0748                                     & 84.48                                     & 0.0486                                      \\
			& LAKERED (\textbf{CVPR 2024})             & $\mathcal{F}$                                                       & 39.55                                     & 0.0212                                     & 88.70                                     & 0.0428                                     & 102.67                                    & 0.0555                                     & 64.27                                     & 0.0355                                      \\
			& Camouflage Anything(\textbf{CVPR 2025})  & $\mathcal{F}$                                                       & \underline{22.30}                                     & \underline{0.0039}                                     & \underline{61.78}                                     & \textbf{0.0211}                                     & \underline{74.53}                                     & \underline{0.0387}                                     & \underline{40.53}                                     & \underline{0.0155}                                      
			\\
			\hline
			\multirow{2}{*}{\begin{tabular}[c]{@{}c@{}}\textit{\textbf{Image }}\\\textit{\textbf{Editing}}\end{tabular}}
			& MIP-Adapter(\textbf{AAAI 2025})          & $\mathcal{I}$ +$\mathcal{T}$+$\mathcal{D}$                                     & 35.32                                     & 0.0265                                     & 99.25                                    & 0.0466                                     & 109.56                                    & 0.0595                                     & 68.26                                     & 0.0391                                      \\
			& {\cellcolor[rgb]{0.925,0.918,0.941}}GenCAMO & {\cellcolor[rgb]{0.925,0.918,0.941}} $\mathcal{I}$ +$\mathcal{T}$+$\mathcal{D}$& {\cellcolor[rgb]{0.925,0.918,0.941}}\textbf{18.49} & {\cellcolor[rgb]{0.925,0.918,0.941}}\textbf{0.0025} & {\cellcolor[rgb]{0.925,0.918,0.941}}\textbf{55.46} & {\cellcolor[rgb]{0.925,0.918,0.941}}\underline{0.0251} & {\cellcolor[rgb]{0.925,0.918,0.941}}\textbf{53.86} & {\cellcolor[rgb]{0.925,0.918,0.941}}\textbf{0.0292} & {\cellcolor[rgb]{0.925,0.918,0.941}}\textbf{38.45} & {\cellcolor[rgb]{0.925,0.918,0.941}}\textbf{0.0123}  \\
			\toprule     % 粗线
	\end{tabular}}
	\label{fid_kid_comparison}
\end{table*}

\paragraph{Diffusion Loss.} 
To enable coherent camouflage generation under multi-conditional guidance, we aim to model the conditional latent distribution 
$z(x \mid C_p, \mathbf{F}_e, \hat{\mathcal{V}}, \mathbf{F}_Q)$. 
To this end, we define a unified diffusion objective that jointly optimizes all multi-modal representations:
\begin{equation}
\begin{array}{c}
\hspace{2.1cm}\underbrace{\hat{\tau}'\leftarrow \mathrm{Fuse}(E_{\text{CLIP}}(C_p), \mathbf{F}_e, \hat{\mathcal{V}})}_{\Downarrow} \\
\hspace{0.1cm}\epsilon_\theta(z_t, t, \hat{\tau}', \mathbf{F}_Q)=\epsilon_\theta(z_t, t, \hat{\tau}')+\mathcal{G}_\phi(\mathbf{F}_Q),
\end{array}
\end{equation}

\begin{equation}
\mathcal{L}_{\text{LDM}}
=
\mathbb{E}_{z,\ \epsilon\sim\mathcal{N}(0,\mathbf{I}),\ t}\!
\left[
\left\|
\epsilon
-
\epsilon_\theta(z_t, t, \hat{\tau}'),\mathbf{F}_Q
\big)
\right\|_{2}^{2}
\right],
\label{eq:gencamo_loss}
\end{equation}

\begin{equation}
\mathcal{L}_{\text{total}}
=
\lambda_{1}\mathcal{L}_{\text{LDM}}
+
\lambda_{2}\mathcal{L}_{\text{DLC}},
\label{eq:total_loss}
\end{equation}
where $\mathrm{Fuse}(\cdot)$ denotes a cross-attention–based modulation function, and both $\lambda_{1}$ and $\lambda_{2}$ are empirically set to $1$ for stable optimization.

\subsection{Synthetic Data Generation}
As shown in Fig.~\ref{fig:model_1}.C, we first derive an environment-aware color from the foreground–background attributes (e.g., a yellow butterfly on green leaves yields green). The controlled image is then fed into GenCAMO to generate the camouflaged result and its latent features. GenCAMO also produces an initial depth map and coarse mask through its depth decoder (trained with an MSE loss) and a DiffuMask-style mask decoder. Finally, Depth Anything and SAM2\cite{ravi2024sam} are used to refine the depth and mask outputs.

%% file: sec/3_finalcopy.tex
\section{Experiment}
\subsection{Experimental Setups}
We evaluate GenCAMO on two tasks: (i) Camouflage Image–Mask Generation (CIG) and (ii) Synthetic-to-Real Camouflage Dense Prediction (S2RCDP), which includes COD, RGB-D COD, and OVCOS. Thanks to the unified construction pipeline, GenCAMO-DB naturally covers all datasets required for these evaluations.

\subsubsection{Datasets}
For the CIG and S2RCDP tasks, we first evaluate on GenCAMO-DB-LAKERED, which includes 4,040 training images and 12,946 testing images from camouflage and salient/general datasets. Under the $\hat{C}2C$ setting \cite{luo2025synthetic}, we use 6,473 synthetic camouflage images to evaluate S2R-COD and S2R-D-COD. For S2R-OVCOS, we train on GenCAMO-DB excluding OVCAMO and LAKERED salient/general data, generate about 3,000 synthetic samples matching the OVCAMO categories, and use them as simulated data for OVCOS training.

\subsubsection{Metrics}
Following \cite{zhao2024lake}, we evaluate CIG with FID \cite{binkowski2018demystifying} and KID \cite{heusel2017gans}. S2RCDP uses MAE, S-measure ($S_m$) \cite{fan2017structure}, E-measure ($E_m$) \cite{fan2018enhanced}, and weighted F-measure ($F_\beta^\omega$) \cite{margolin2014evaluate}. For OVCOS, we employ task-adapted metrics, $cS_m$, $cF_w^\beta$, $cMAE$, and $cE_m$, following OVSIS conventions \cite{cho2024cat, luo2025synthetic} to capture both reasoning and segmentation performance.

\subsubsection{Implementation Details}
We build our reference text-to-image framework on Stable Diffusion v1.5 with ControlNet and OpenCLIP ViT-H/14 as the image encoder. For generation, we compare against LAKE-RED and MIP-Adapter, and for downstream evaluation, we use SINet/SINet-v2 (S2R-COD), RISNet (S2R-D-COD), and OVCoser (S2R-OVCOS). We additionally adopt CSRDA\cite{luo2025synthetic}, an unsupervised domain adaptation strategy for aligning synthetic and real data in S2R tasks.

\begin{figure*}[t]
	\centering
	\includegraphics[width=\linewidth]{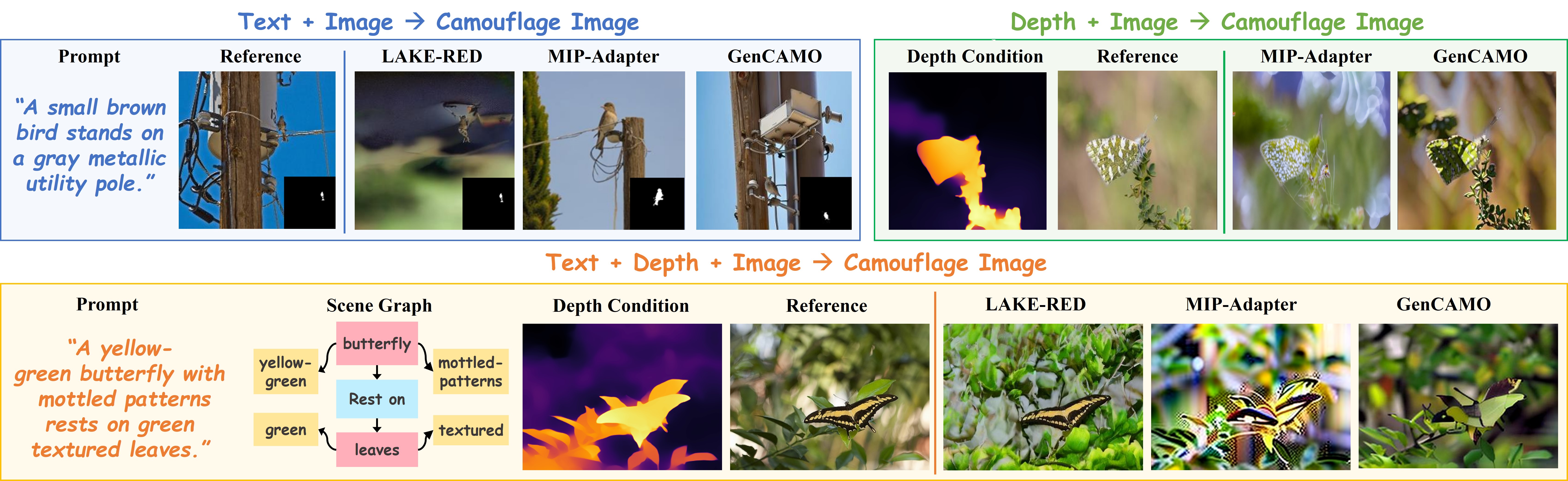}
	\caption{Multi-modal controllable camouflage image synthesis. Comparison of LAKE-RED, MIP-Adapter, and GenCAMO under Text + Image, Depth + Image, and Text + Depth + Image with the scene graph.}
	\label{fig_cvpr}
\end{figure*}

\subsection{Comparison of generation}

\subsubsection{Quantitative Comparison.}
As shown in Tab.~\ref{fid_kid_comparison}, our method achieves the best overall FID and KID scores, surpassing all baselines. The gains are most notable on the challenging ``General Objects” category, reflecting the stronger generalization and semantic reasoning brought by our multi-modal design.

\begin{table}
		\caption{Experimental results of S2R-COD and S2R-D-COD task on CAMO + NC4K + CHAMELEON → COD10K ($\hat{C}2C$) benchmark.}
	\centering
	\renewcommand{\arraystretch}{1.3}
	\resizebox{\linewidth}{!}{
		\begin{tabular}{c|c|cccc} 
			\toprule     % 粗线
			Model                     & Setting      & $S_m$\textbf{$\uparrow$} & $F_w^\beta$\textbf{$\uparrow$} & $E_m$\textbf{$\uparrow$}   & MAE\textbf{$\downarrow$}      \\ 
			\hline
			
			\multirow{3}{*}{\textbf{\textit{SINet + CSRDA}}}    
			& LAKE-RED     & \underline{0.7555}     & 0.5202     & 0.7779  & 0.0650  \\ 
			& MIP-Adapter     & 0.7458     & \underline{0.5282}     & \underline{0.7865}  & \underline{0.0645}  \\ 
			\cline{2-6}
			& {\cellcolor[rgb]{0.925,0.918,0.941}}Ours         
			& {\cellcolor[rgb]{0.925,0.918,0.941}}\textbf{0.7818}     
			& {\cellcolor[rgb]{0.925,0.918,0.941}}\textbf{0.5983}     
			& {\cellcolor[rgb]{0.925,0.918,0.941}}\textbf{0.8076}  
			& {\cellcolor[rgb]{0.925,0.918,0.941}}\textbf{0.0460}  \\ 
			\hline
			
			\multirow{3}{*}{\textbf{\textit{SINet-v2 + CSRDA}}} 
			& LAKE-RED     & 0.721     & 0.5329     & \underline{0.7975}  & 0.0656  \\
			& MIP-Adapter     & \underline{0.7303}     & \underline{0.5396}     & 0.7869  & \underline{0.0649}  \\  
			\cline{2-6}
			& {\cellcolor[rgb]{0.925,0.918,0.941}}Ours         
			& {\cellcolor[rgb]{0.925,0.918,0.941}}\textbf{0.7874}     
			& {\cellcolor[rgb]{0.918,0.918,0.941}}\textbf{0.6338}     
			& {\cellcolor[rgb]{0.918,0.918,0.941}}\textbf{0.8622}  
			& {\cellcolor[rgb]{0.918,0.918,0.941}}\textbf{0.0431}  \\ 
			\hline
			
			\multirow{3}{*}{\textbf{\textit{RISNet + CSRDA}}}   
			& LAKE-RED     & 0.7745     & \underline{0.6157}     & \underline{0.8334}  & \underline{0.0519}  \\ 
			& MIP-Adapter     & \underline{0.7796}     & 0.6134     & 0.8299  & 0.0525  \\ 
			\cline{2-6} 
			&  {\cellcolor[rgb]{0.925,0.918,0.941}} Ours         
			& {\cellcolor[rgb]{0.925,0.918,0.941}}\textbf{0.8036}     
			& {\cellcolor[rgb]{0.925,0.918,0.941}}\textbf{0.6645}     
			& {\cellcolor[rgb]{0.925,0.918,0.941}}\textbf{0.8675}  
			& {\cellcolor[rgb]{0.925,0.918,0.941}}\textbf{0.0423}  \\
			\toprule     % 粗线
	\end{tabular}}
	\label{cod_table}
\end{table}

\subsubsection{Qualitative Comparison.}
As shown in Fig.~\ref{fig_cvpr}, We compare LAKE-RED, MIP-Adapter, and GenCAMO under multiple condition settings. GenCAMO achieves stronger semantic alignment, geometric consistency, and appearance transfer. Depth cues stabilize scale and occlusion, while scene-graph guidance improves object–context relations. Overall, GenCAMO yields more natural blending and illumination consistency, resulting in stronger and more controllable camouflage.

\subsection{Comparison in dense prediction}

\subsubsection{Quantitative Comparison.}

As shown in Tab.~\ref{cod_table}, adding our synthetic data yields consistently better COD performance than LAKE-RED or MIP-Adapter. Our samples reduce the synthetic–real gap more effectively, enabling models to learn clearer and more reasoning camouflage cues. Likewise, Fig.~\ref{table_ovcamo} shows that OV-Camo trained with GenCAMO data—alone or combined with real images—achieves higher accuracy than real-only training. GenCAMO alone is competitive, and the combined setting performs best, indicating that our generated data further strengthens model training.

\begin{table}
	\caption{Quantitative results of S2R-OVCOS using the OV-Camo model under different training settings.}
	\centering
	\renewcommand{\arraystretch}{1.3}
	\resizebox{\linewidth}{!}{
		\begin{tabular}{c|c|c|cccc} 
			\toprule
			\multirow{2}{*}{Model}  
			& \multicolumn{2}{c|}{Training Data} 
			& \multirow{2}{*}{$cS_m$\textbf{$\uparrow$}} 
			& \multirow{2}{*}{$cF_w^\beta$\textbf{$\uparrow$}} 
			& \multirow{2}{*}{cMAE\textbf{$\downarrow$}} 
			& \multirow{2}{*}{$cE_m$\textbf{$\uparrow$}} 
			\\ 
			\cline{2-3}
			& Real & GenCamo & & & & \\ 
			\hline
			
			\multirow{3}{*}{\textbf{\textit{OVCamo}}} 
			& $\times$    & $\checkmark$                                                         
			& \underline{0.579} 
			& \underline{0.490}
			& \underline{0.336}
			& \underline{0.616}
			\\
			
			& $\checkmark$ & $\times$                                      
			& 0.547 
			& 0.442
			& 0.394
			& 0.579
			\\
			
			& $\checkmark$  & $\checkmark$ 
			& {\cellcolor[rgb]{0.925,0.918,0.941}}\textbf{0.589}
			& {\cellcolor[rgb]{0.925,0.918,0.941}}\textbf{0.518}
			& {\cellcolor[rgb]{0.925,0.918,0.941}}\textbf{0.311}
			& {\cellcolor[rgb]{0.925,0.918,0.941}}\textbf{0.657}
			\\
			\toprule
	\end{tabular}}
	\label{table_ovcamo}
\end{table}

\begin{table}
		\caption{Quantitative ablation results under different module settings.}
	\centering
	\resizebox{0.5\linewidth}{!}{
	\renewcommand{\arraystretch}{1.3}
	\begin{tabular}{cc|cc} 
			\toprule     % 粗线
		\multicolumn{2}{c|}{Modules} & \multicolumn{2}{c}{Overall}        \\ 
		\hline
		DLCG        & AMA           & FID$\downarrow$ & KID$\downarrow$  \\ 
		\hline
		$\times$     & $\times$      & 54.32           & 0.0239           \\
		$\times$     & $\checkmark$  & 43.45           & \underline{0.0172}   \\
		$\checkmark$ & $\times$      & \underline{42.57}   & 0.0192           \\
		$\checkmark$ & $\checkmark$  & {\cellcolor[rgb]{0.933,0.918,0.949}}\textbf{38.45}  & {\cellcolor[rgb]{0.933,0.918,0.949}}\textbf{0.0123}  \\
			\toprule     % 粗线
	\end{tabular}}
	\label{table1_ablative}
\end{table}

\begin{figure}
	\centering
	\includegraphics[width=0.48\textwidth]{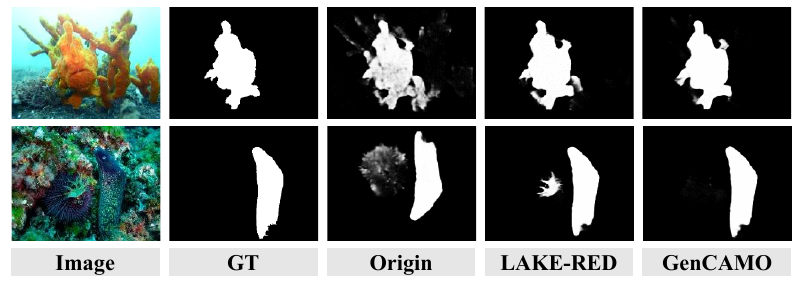} % Reduce the figure size so that it is slightly narrower than the column.
	\caption{Qualitative comparison of concealed object segmentation results on COD10K using RISNet + CSRDA under S2R-D-COD setting.}
	\label{fig_cod}
\end{figure}

\begin{figure*}
	\centering
	\includegraphics[width=\linewidth]{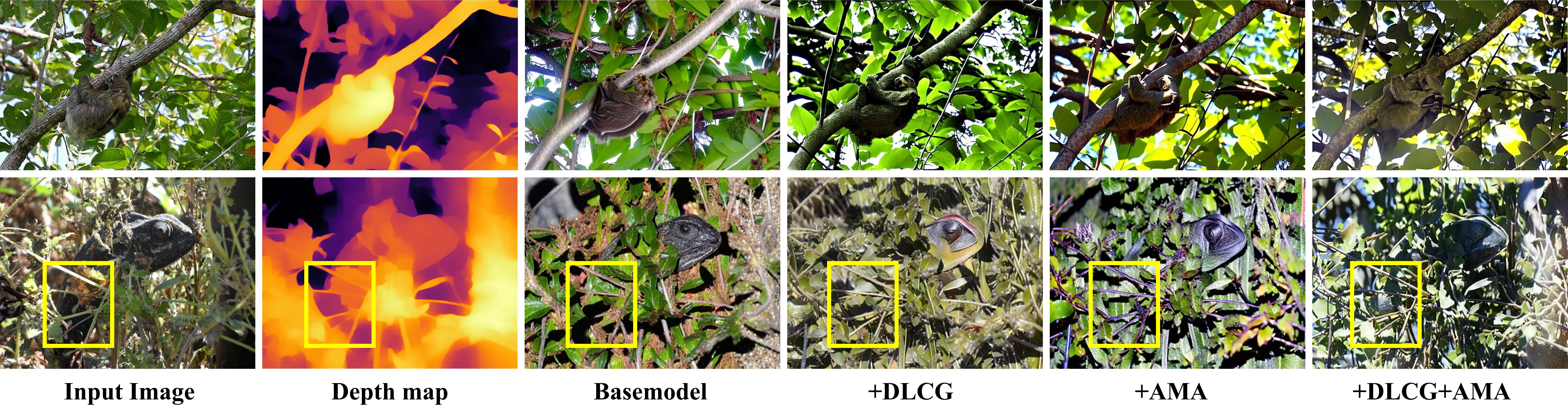} 
	\caption{Qualitative ablation on camouflaged object generation. From left to right: Input Image, Depth map, Basemodel, +DLCG, +AMA, and +DLCG+AMA.}
	\label{fig_ablative}
\end{figure*}

\begin{figure}
	\centering
	\includegraphics[width=\linewidth]{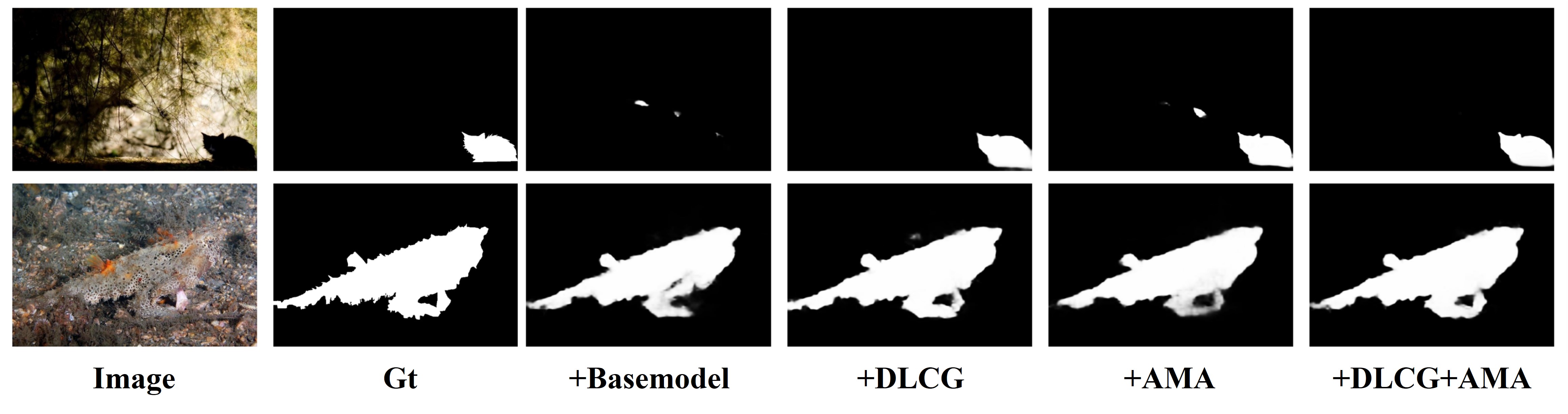} 
	\caption{Qualitative comparison of camouflage image segmentation under different module settings.}
	\label{fig_cod_ablative}
\end{figure}

\subsubsection{Qualitative Comparison.}

As shown in Fig.~\ref{fig_cod}, models trained without synthetic data or with LAKE-RED still struggle to produce stable camouflage predictions. In contrast, incorporating our GenCAMO synthetic data leads to noticeably more coherent and reliable dense prediction results. This improvement demonstrates that GenCAMO provides stronger supervision for S2R-COD and S2R-D-COD.

\subsection{Ablative Study}

\subsubsection{Quantitative Ablation.}
As shown in Tab.~\ref{table1_ablative}, introducing either DLCG or AMA brings clear gains in both FID and KID. DLCG yields the largest FID improvement (reducing error by over 20\%), reflecting stronger depth–layout coherence, while AMA achieves the largest KID improvement (a nearly 30\% reduction), indicating better attribute-level alignment. When combined, the two modules produce an additional 10–15\% overall gain, achieving the best results on both metrics. These improvements highlight the complementary strengths of DLCG and AMA: DLCG enhances geometric structure, AMA refines appearance consistency, and together they lead to a more stable synthesis distribution that benefits downstream segmentation.

\subsubsection{Qualitative Ablation.}

As shown in Fig.~\ref{fig_ablative}, the Base model exhibits oversmoothed textures and unclear boundaries, especially in the yellow-highlighted occluded regions. Incorporating DLCG improves geometric plausibility through depth-guided cues, while AMA enhances local appearance consistency. With both modules, the scene-graph–enhanced model recovers finer object details under occlusion and achieves more coherent foreground–background blending.
Consistently, Fig.~\ref{fig_cod_ablative} further confirms that combining DLCG and AMA yields the most accurate segmentation masks, demonstrating the effectiveness of our scene-graph–enhanced modeling under occlusion.

\begin{figure}
	\centering
	\includegraphics[width=\linewidth]{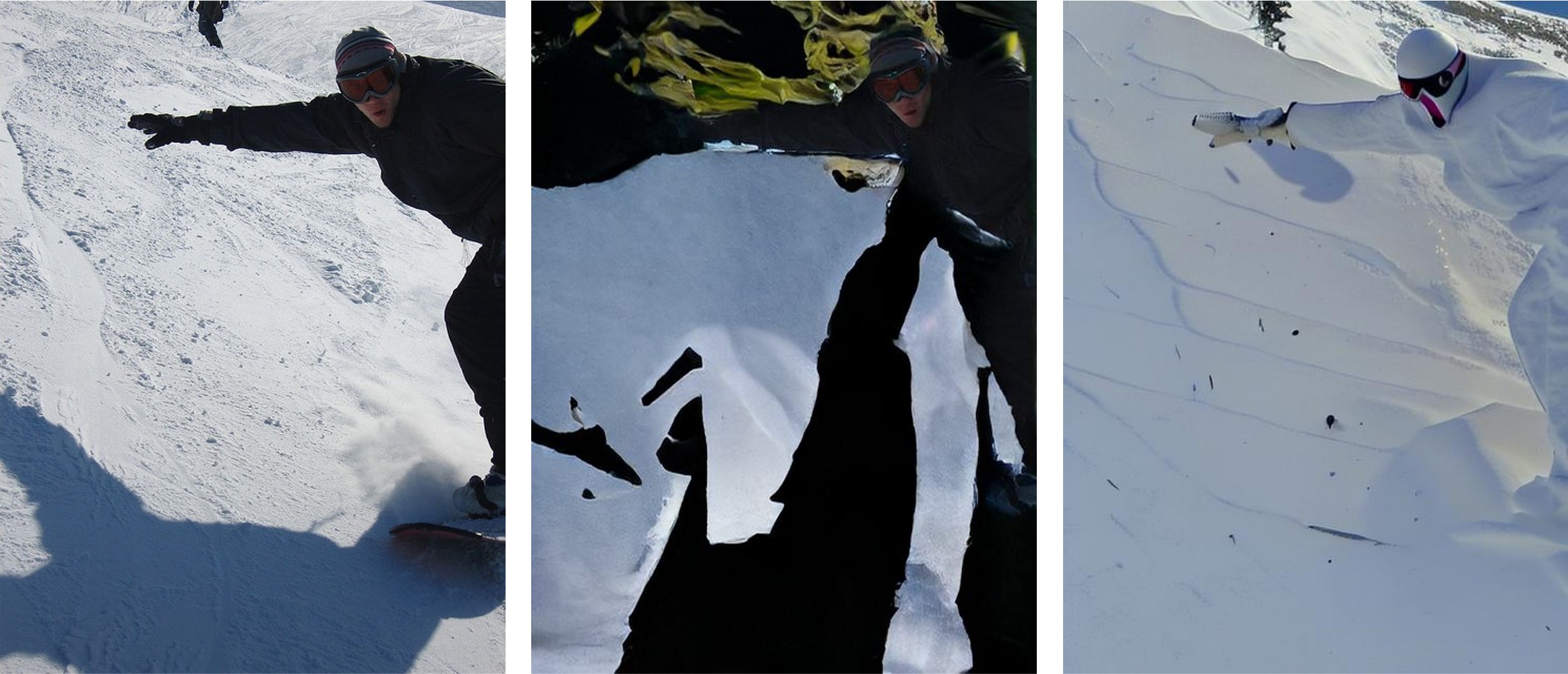} 
	\caption{Failure cases. Input image, LAKE-RED result, and GenCAMO result (left to right). GenCAMO achieves global camouflage, but some local details remain insufficiently concealed, such as the red face covering.}
	\label{failure}
\end{figure}

\subsubsection{Limitation and Future Improvement.} Although our method generates convincing camouflage effects, two limitations remain. Local appearance cues may still cause artifacts (e.g., dark-red goggles rendered as a solid mask; Fig.~\ref{failure}), and the model has difficulty handling realistic illumination and shadows. In future work, we will explore finer feature alignment and physics-aware priors to further enhance visual fidelity and robustness.

\section{Conclusion}
This paper investigates reference-guided text-to-image diffusion modeling for generate camouflage image–dense annotations without requiring entensive manually annotated foreground masks, enabling more robust training of concealment-related dense prediction models across diverse camouflage scenes.
To support this goal, we curate GenCAMO-DB, a large-scale camouflage image–text dataset enriched with multiple metadatas, including fine-grained attribute descriptions, depth maps, scene graphs. 
Built upon this dataset, we introduce GenCAMO, an environment-aware and mask-free generative framework capable of synthesizing high-fidelity camouflage images together with dense annotations. 
Extensive experiments across various synthetic-to-real camouflage dense prediction tasks verify that GenCAMO significantly enhances the robustness of camouflage scene understanding models, especially in unannotated or mask-scarce condition.
%We believe this work offers a promising direction for mitigating camouflage dense-data scarcity and advancing camouflage perception and generation.

%% file: sec/X_suppl.tex
\clearpage
\appendix
\setcounter{page}{1}
\maketitlesupplementary

\section{More Analysis of GenCAMO-DB}
\noindent
\textbf{Text.} To obtain rich textual descriptions that reflect camouflage-related semantics, we design a structured prompt for GPT4o\cite{hurst2024gpt} that explicitly guides the model to describe each image using object attributes, object categories, and inter-object relations. 
Specifically, the prompt instructs large language model (LLM) to generate a comprehensive sentence following a subject–verb–object (SVO) pattern, where both the subject and object are enriched with modifiers describing their colors, textures, and other appearance cues. This design ensures that the resulting text representations provide comprehensive attribute, object, and relation information aligned with the requirements of scene-graph construction.

\noindent
\begin{quote}
	\colorbox{gray!15}{\parbox{\linewidth}{
			\textit{
				``Describe the image in one concise sentence.  
				Use a subject--verb--object structure to state what the animal is doing.  
				Modify both the subject and object with color, texture, and appearance descriptors.  
				Include concealment cues describing how the animal blends with its surroundings.  
				Add environment cues that specify the background materials or habitats.  
				Explicitly mention spatial or contact relations (e.g., lies on, hides in, blends with).''}
	}}
\end{quote}

\noindent
\textbf{Depth.} The depth contrast distribution exhibits a clear unimodal pattern centered around moderate contrast values, indicating that in most scenes, foreground objects and their surrounding backgrounds maintain similar depth levels. 
This reflects the geometric nature of camouflage in real environments, where organisms typically remain close to surfaces such as leaves, branches, ground, or rocks to minimize depth discontinuities.

\begin{figure}[t]
	\centering
	\includegraphics[width=\linewidth]{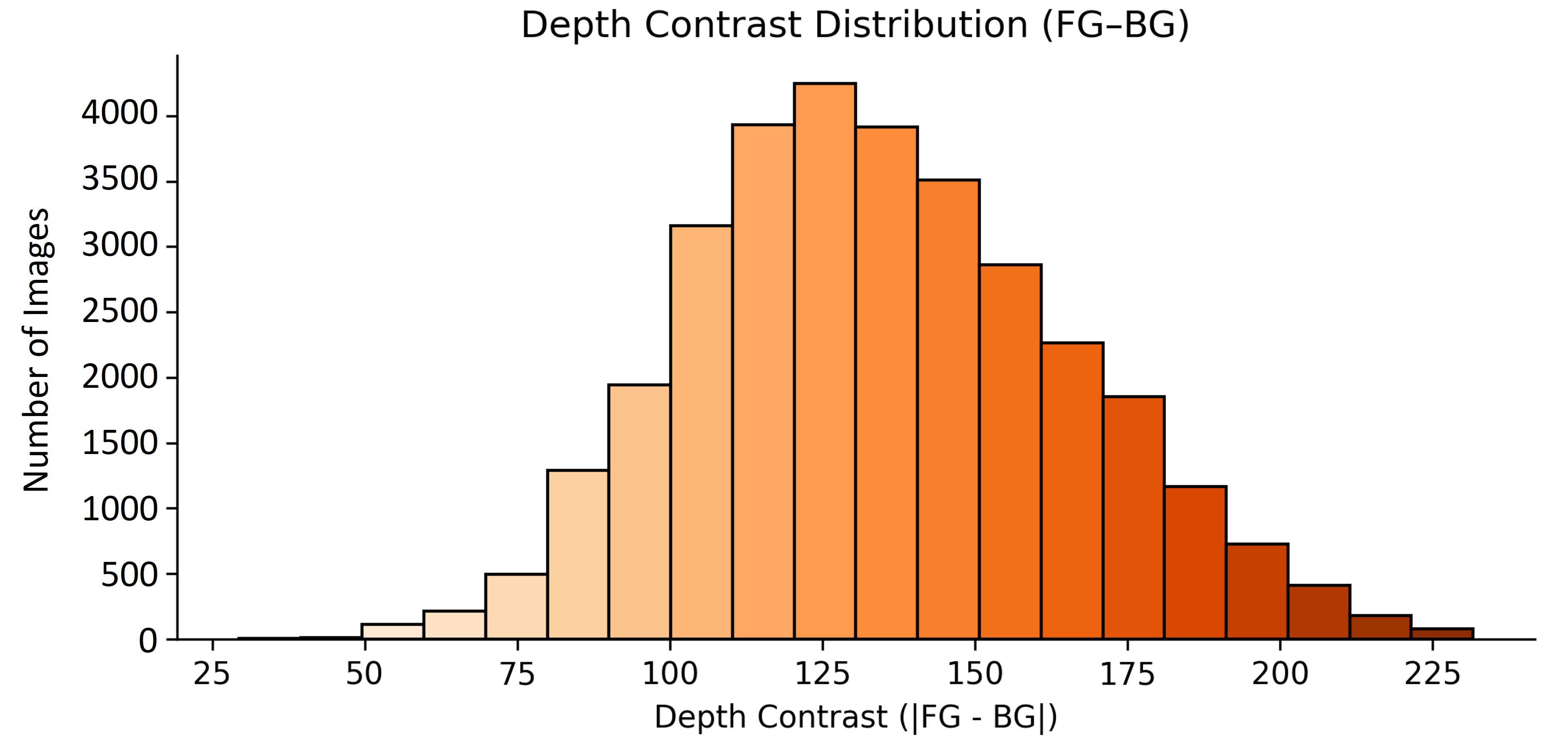}
	\caption{Histogram of foreground–background depth contrast computed from GenCAMO-DB, showing the distribution of depth differences across all samples.}
	\label{appendix_depth}
\end{figure}

As shown in Fig.\ref{appendix_depth}, the long tail toward lower contrast confirms the presence of hard geometric-camouflage cases, where foreground and background depths are nearly identical, increasing scene ambiguity. Meanwhile, the tail toward higher contrast corresponds to easier cases, where foreground objects stand out due to noticeable geometric separation.

Overall, the distribution demonstrates that GenCAMO-DB offers a balanced spectrum of easy-to-hard geometric camouflage conditions, ensuring that models trained on this dataset can learn robust depth-aware camouflage reasoning rather than relying solely on RGB appearance cues.

\begin{figure}[t]
	\centering
	\includegraphics[width=\linewidth]{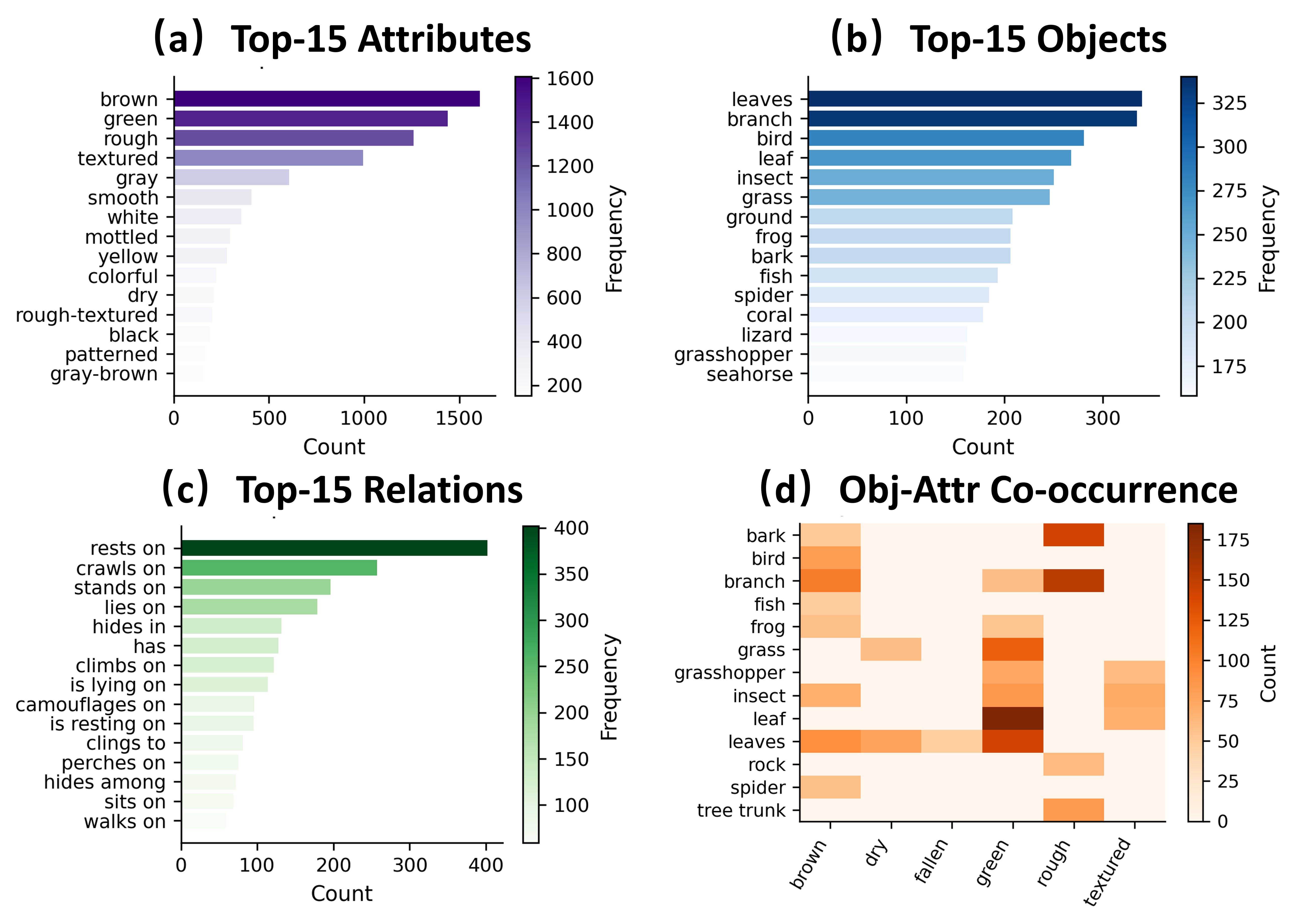}
	\caption{Top-15 distributions of attributes, objects, relations, and object–attribute co-occurrences extracted from GenCAMO-DB’s scene-graph annotations.}
	\label{appendix_fig_scene_graph}
\end{figure}

\noindent
\textbf{Scene Graph.}
Fig.\ref{appendix_fig_scene_graph} illustrates that the scene-graph annotations in GenCAMO-DB strongly emphasize the key factors of camouflage.
On the attribute side, the Top-15 attributes are dominated by colors and textures characteristic of natural concealment (e.g., brown, green, rough, textured), revealing strong appearance similarity between foreground organisms and their backgrounds.
On the relation side, the most frequent relations (e.g., rests on, crawls on, lies on, hides in) describe close physical contact and contextual attachment between objects and the environment.
This demonstrates that GenCAMO-DB captures not only appearance-level camouflage cues but also context- and geometry-level camouflage behaviors, enabling scene-graph–driven generation to reason about both visual similarity and spatial embedding.

%\section{More De/tails of GenCAMO}

\section{Preliminaries of Conditional Text-to-Image Diffusion Models}

Diffusion models (DMs)\cite{song2020denoising} are a class of generative models that learn the data distribution $p(x)$ by gradually denoising a noisy variable $x_T$ sampled from a Gaussian prior $\mathcal{N}(0, \mathbf{I})$.
Their training can be viewed as learning the reverse process of a fixed-length Markov chain consisting of $T$ denoising steps. 
To generate high-resolution images efficiently, Latent Diffusion Models (LDMs)\cite{rombach2022high} encode the image $x$ into a latent representation $z$ using a pretrained autoencoder, and learn the distribution $p(z)$ instead of $p(x)$.
For text-to-image generation, the text prompt condition $C_p$ is first embedded by a frozen CLIP\cite{radford2021learning} text encoder $E_{\text{CLIP}}(C_p)$, and the diffusion model learns to predict the added noise $\epsilon$ through a denoising objective:
\begin{equation}
\mathcal{L}_{\text{T2I}} =
\mathbb{E}_{z,\, \epsilon \sim \mathcal{N}(0,\mathbf{I}),\, t}
\left[
\left\| \epsilon - \epsilon_\theta(z_t, E_{\text{CLIP}}(C_p), t) \right\|_2^2
\right],
\end{equation}
where $t$ is a randomly sampled diffusion step, $\epsilon_\theta$
denotes the noise prediction network with learnable parameters $\theta$.

Based on the standard text-conditioned LDM objective, we further extend the model to a multi-conditional formulation for controllable and high-quality camouflage image generation with dense annotations, as follows:
\begin{equation}
\begin{array}{c}
\hspace{2.1cm}\underbrace{\hat{\tau}\leftarrow \mathrm{Fuse}(E_{\text{CLIP}}(C_p), C_r)}_{\Downarrow} \\
\hspace{0.1cm}\epsilon_\theta(z_t, t, \hat{\tau}, C_d)=\epsilon_\theta(z_t, t, \hat{\tau})+\mathcal{G}_\phi(C_d),
\end{array}
\end{equation}
where $C_r$ indicates the reference image, $\hat{\tau}$ represents the visual–text feature obtained through the cross-attention modulation function $ \mathrm{Fuse}(\cdot)$, and $\mathcal{G}_\phi$ denotes the ControlNet\cite{zhang2023adding} module parameterized by $\phi$, which provides structural guidance conditioned on the depth input $C_d$.

%\subsection{Image-Dense Annotation Generation}
%xxxx

\begin{figure}[t]
	\centering
	\includegraphics[width=\linewidth]{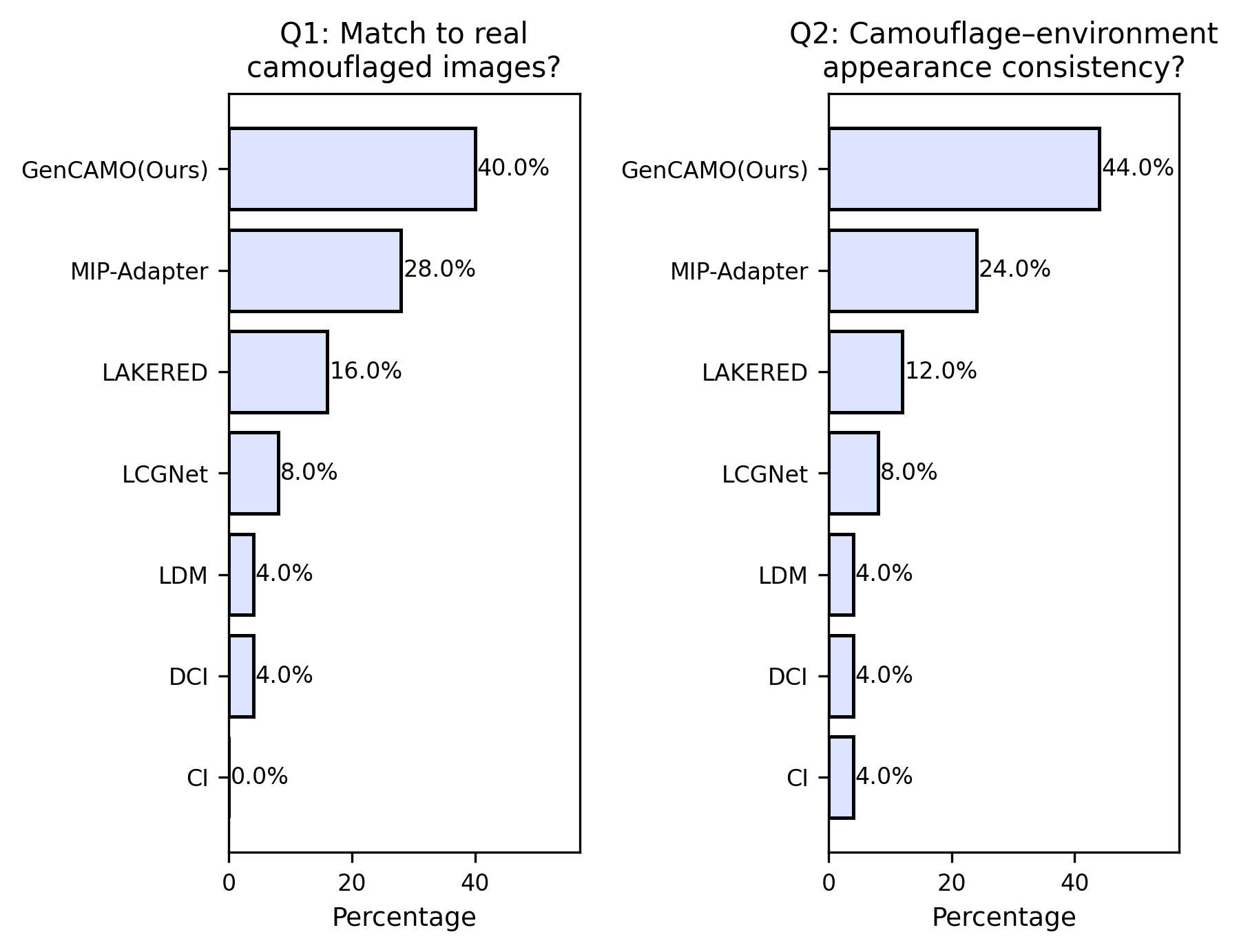}
	\caption{Results of the user study evaluating subjective judgments of camouflaged image generation across different methods.
		GenCAMO receives the highest preference in both realism matching and camouflage–environment consistency, indicating that it produces results most aligned with real-world camouflage perception.}
	\label{appendix_user_study}
\end{figure}

\subsection{More Examples from Synthetic Camouflaged Dataset}

\begin{figure*}[t]
	\centering
	\includegraphics[width=\linewidth]{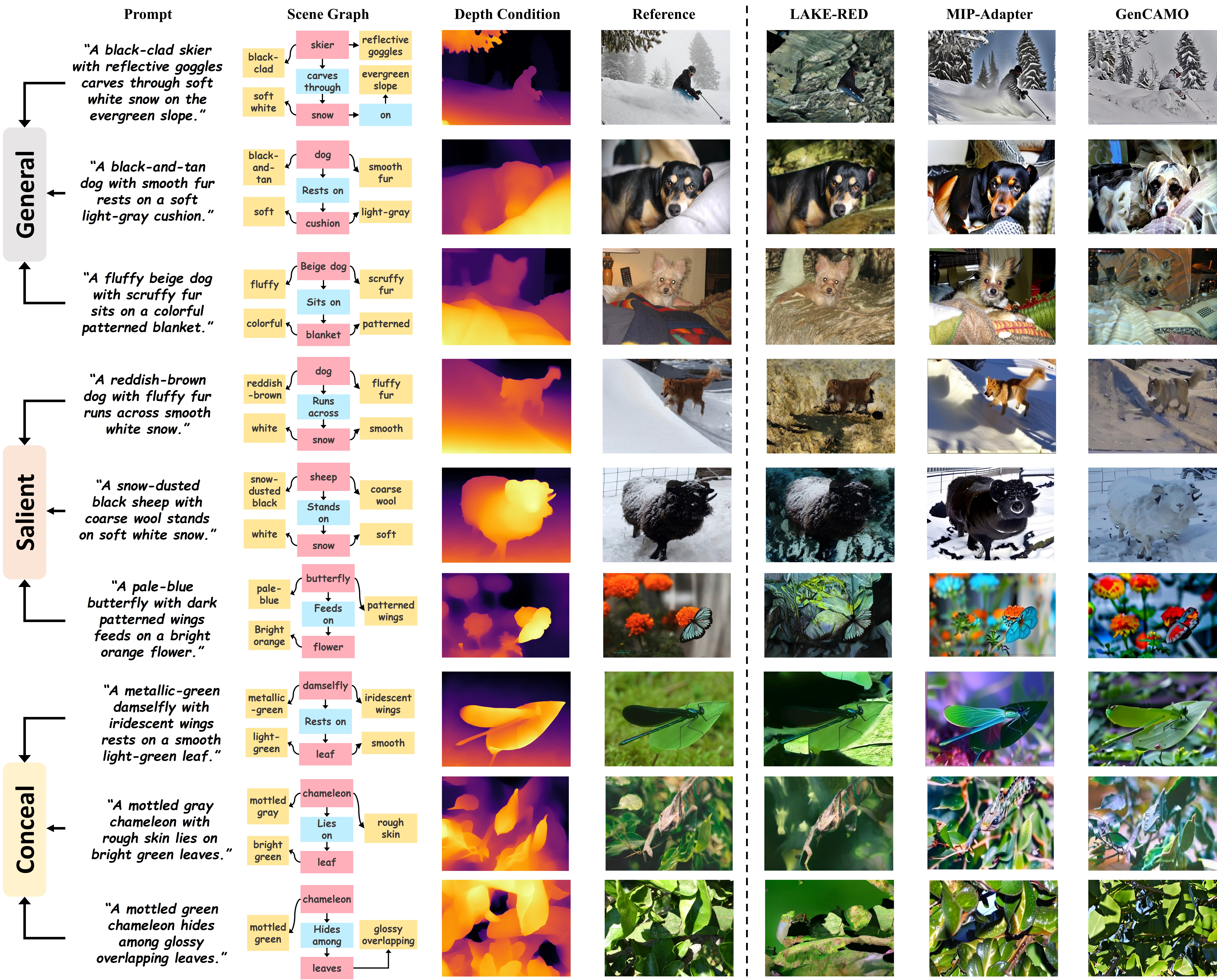}
	\caption{Qualitative comparison across \textit{general}, \textit{salient}, and \textit{conceal} cases. 
		Given the prompt, scene graph, and depth condition derived from GenCAMO-DB, our GenCAMO produces visually coherent and environment-aware camouflage results, compared with LAKE-RED and MIP-Adapter..}
	\label{appendix_fig_generate_image}
\end{figure*}

\section{Additional Experimental Results}
\subsection{User Study}

As the visual quality of camouflage generation is inherently tied to human perception, 
we conducted a user study to collect subjective evaluations of the synthesized results. 
Following the standard practice for perceptual evaluation in camouflage generation\cite{zhao2024lake}, 
we randomly sampled 100 images from each of the three subsets of our GenCAMO-DB dataset 
(COD, SOD, and SEG), resulting in a total of 300 images.

To ensure a comprehensive evaluation, our GenCAMO framework was compared with a wide range of representative camouflage-generation approaches, including CI~\cite{chu2010camouflage}, DCI~\cite{zhang2020deep}, LDM~\cite{rombach2022high}, LCGNet~\cite{chen2025foreground}, LAKE-RED~\cite{zhao2024lake}, and MIP-Adapter~\cite{huang2025resolving}. 
All competing methods were applied to generate their corresponding camouflaged outputs under the same experimental protocol.
For style-transfer-based approaches, such as CI, DCI and LCGNet, an auxiliary background image was uniformly sampled 
from the Places365\cite{zhou2017places} dataset and kept identical across all methods to ensure a fair comparison. 
These generated results were then shown to 25 human participants, 
who were asked to provide subjective judgments based on two key questions 
designed to reflect the core objectives of camouflage generation, 
namely visual realism and camouflage--environment appearance consistency:

\begin{itemize}
	\item (\textbf{Q1})~Which result best matches real camouflaged images observed in real-world scenes?
	\item (\textbf{Q2})~Which method achieves the strongest appearance consistency between the camouflage object and its surrounding environment?
\end{itemize}

For each question, participants selected their top three preferred results, with rank~1 indicating the strongest preference. 
The aggregated voting outcomes are presented in Fig.~\ref{appendix_user_study}. 
Across both evaluation aspects, \textbf{GenCAMO receives the highest proportion of votes}, 
surpassing all competing approaches by a clear margin.  
While several baselines may occasionally produce visually plausible results, they typically fail to maintain coherent environmental adaptation or realistic appearance blending.  
In contrast, GenCAMO consistently generates images perceived as both 
(\textit{i}) closest to real-world camouflage examples and 
(\textit{ii}) most consistent with the surrounding environment, 
verifying the effectiveness of our environment-aware camouflage generation framework.

Figure~\ref{appendix_fig_generate_image} presents qualitative comparisons across three representative categories in GenCAMO-DB, 
including \textit{general}, \textit{salient}, and \textit{conceal} scenarios. 
Although LAKE-RED is able to produce visually camouflaged patterns in several examples by expanding background textures 
(e.g., the sheep and butterfly cases), the outpainting nature of this pipeline often leads to geometric distortions and 
inconsistent object boundaries. In contrast, generation-based methods such as MIP-Adapter produce results with higher image 
realism and scene-level consistency.

Building on explicit scene-graph decoupling, our GenCAMO further achieves accurate object--environment integration, resolving the inherent limitations of reference-guided conditional text-to-image models in camouflage generation. For instance, in the skier example, GenCAMO preserves both the foreground geometry and the compatibility between snow textures and illumination, whereas LAKE-RED produces an overly blended silhouette that deviates from realistic camouflage. MIP-Adapter generates visually plausible appearances, but often struggles with foreground fidelity and scene semantics. 

In the snow–sheep and snow–dog examples, MIP-Adapter introduces unintended auxiliary objects and background artifacts, while GenCAMO maintains correct foreground structure and produces consistent snow–fur camouflage cues. Similarly, in the butterfly case, MIP-Adapter suffers from color confusion between the insect and surrounding flowers, leading to ambiguous object boundaries; GenCAMO instead aligns the object's color, texture, and spatial relations with the environment, resulting in clearer yet naturally concealed patterns. Overall, GenCAMO achieves stronger scene-aware camouflage generation than both LAKE-RED and MIP-Adapter.

Overall, our results demonstrate that GenCAMO can reliably synthesize high-fidelity camouflage 
images across diverse visual scenarios, including general objects, salient targets, and 
challenging concealment cases. This broad applicability enables the generation of otherwise 
difficult camouflage samples that are rarely captured in real-world datasets. Consequently, 
GenCAMO offers a scalable solution for enriching data in multiple image–dense prediction tasks, 
effectively alleviating data scarcity and improving downstream model robustness across 
camouflage-intensive environments.

%\section{Rationale}
%\label{sec:rationale}
%% 
%Having the supplementary compiled together with the main paper means that:
%% 
%\begin{itemize}
%\item The supplementary can back-reference sections of the main paper, for example, we can refer to \cref{sec:intro};
%\item The main paper can forward reference sub-sections within the supplementary explicitly (e.g. referring to a particular experiment); 
%\item When submitted to arXiv, the supplementary will already included at the end of the paper.
%\end{itemize}
%% 
%To split the supplementary pages from the main paper, you can use \href{https://support.apple.com/en-ca/guide/preview/prvw11793/mac#:~:text=Delete%20a%20page%20from%20a,or%20choose%20Edit%20%3E%20Delete).}{Preview (on macOS)}, \href{https://www.adobe.com/acrobat/how-to/delete-pages-from-pdf.html#:~:text=Choose%20%E2%80%9CTools%E2%80%9D%20%3E%20%E2%80%9COrganize,or%20pages%20from%20the%20file.}{Adobe Acrobat} (on all OSs), as well as \href{https://superuser.com/questions/517986/is-it-possible-to-delete-some-pages-of-a-pdf-document}{command line tools}.